\documentclass[conference]{IEEEtran}
\usepackage{cite}
\usepackage{amsmath,amssymb,amsfonts}
\usepackage{algorithmic}
\usepackage{graphicx}
\usepackage{textcomp}
\usepackage{xcolor}
\usepackage{float}
\usepackage{url}
\def\BibTeX{{\rm B\kern-.05em{\sc i\kern-.025em b}\kern-.08em
    T\kern-.1667em\lower.7ex\hbox{E}\kern-.125emX}}
\begin{document}

\title{Evaluation of Machine and Deep Learning Techniques for Cyclone Trajectory Regression and Status Classification by Time Series Data\\
\thanks{}
}

\author{\IEEEauthorblockN{Ethan Zachary Lo}
\IEEEauthorblockA{\textit{}\\
\textit{Walton High School}\\
Marietta, GA, USA \\
elo9908@gmail.com}
\and
\IEEEauthorblockN{Dan Chia-Tien Lo}
\IEEEauthorblockA{\textit{Department of Computer Science} \\
\textit{Kennesaw State University}\\
Marietta, GA, USA \\
dlo2@kennesaw.edu}
}

\maketitle

\begin{abstract}
Accurate cyclone forecasting is essential for minimizing loss of life, infrastructure damage, and economic disruption. Traditional numerical weather prediction models, though effective, are computationally intensive and prone to error due to the chaotic nature of atmospheric systems. This study proposes a machine learning (ML) approach to forecasting tropical cyclone trajectory and status using time series data from the National Hurricane Center, including recently added best track wind radii. A two-stage ML pipeline is developed: a regression model first predicts cyclone features—maximum wind speed, minimum pressure, trajectory length, and directional change—using a sliding window of historical data. These outputs are then input into classification models to predict the cyclone’s categorical status. Gradient boosting regression and three classifiers—random forest (RF), support vector machine (SVM), and multi-layer perceptron (MLP)—are evaluated. After hyperparameter tuning and synthetic minority oversampling (SMOTE), the RF classifier achieves the highest performance with 93\% accuracy, outperforming SVM and MLP across precision, recall, and F1 score. The RF model is particularly robust in identifying minority cyclone statuses and minimizing false negatives. Regression results yield low mean absolute errors, with pressure and wind predictions within ~2.2 mb and ~2.4 kt, respectively. These findings demonstrate that ML models, especially ensemble-based classifiers, offer an effective, scalable alternative to traditional forecasting methods, with potential for real-time cyclone prediction and integration into decision-support systems.

\end{abstract}


\section{Introduction}
The National Weather Service defines a tropical cyclone as a rotating, organized system of storms that forms over tropical or subtropical waters, characterized by a closed low-level circulation \cite{NWS2024}. According to the United Nations, the U.S. experiences the highest annual number of tropical cyclone impacts, averaging 12 per year \cite{UNDP2004}. The frequent occurrence of cyclone impacts in the U.S. has exacerbated their impacts on the environment and the economy. Since 1980, of the 291 billion-dollar weather and climate disasters that have occurred in the U.S. alone, the most costly catastrophes have been tropical storms and hurricanes, amounting to nearly \$1 trillion in damages \cite{AMS2021}. From 1980 to 2022, tropical cyclones accounted for approximately 7,211 deaths in the U.S., making them one of the deadliest weather phenomena \cite{NCEI2025}. Despite advancements, humans cannot physically prevent these catastrophic events. Humans have therefore adapted to depend on forecasted predictions, which is why there is a need for precise and efficient prediction models to mitigate their impact. 

Traditional weather prediction models perform complex mathematical and physical calculations from atmospheric data to provide information for research insights and regional forecasts. Because commutes or general routines can be impacted by the weather, this forecasted information is used daily. With rapidly evolving phenomena, such as thunderstorms, cyclones, hurricanes, etc., accurate preemptive data is essential for government agencies and individuals to make timely, coordinated responses in emergencies \cite{NOAA2024}. Slightly incorrect predictions like underestimating wind speed can result in shortsighted decisions, resulting in exponentially higher damages and post-recovery spending \cite{MolinaRudik2022}. Furthermore, small variations in interrelated variables, as described by chaos theory \cite{Biswas2018}, can significantly alter forecasts. These variables cannot be controlled, but they can be predicted with precision.

To account for uncertainties, meteorologists employ ensemble methods, which are techniques that use multiple models performed on different parameters. Executing large, complex calculations with high-resolution data additionally consumes extensive computational power \cite{NOAAHurricaneModel}. Advanced forecasting models, such as the European Centre for Medium-Range Weather Forecasts (ECMWF) and the High-Resolution Rapid Refresh (HRRR), require supercomputers capable of performing up to 12 quadrillion calculations per second \cite{IBM2025}. 

Machine learning (ML) models have significantly contributed to solving complex problems in various fields such as healthcare, finance, and artificial intelligence. As an alternative, ML models can be an effective solution to the high computational power and runtime during predictions, as trained models can execute instantly without significant variances in accuracy. In this paper, several ML models are implemented on cyclone data to examine their performance and results. The tested models include gradient boosting regression (GBR), random forest (RF) classification, support vector machine (SVM) classification, and multi-layer perceptron (MLP) classification, all imported from scikit-learn \cite{ScikitLearn}, a popular Python library for machine learning. The training variables include general cyclone features in addition to best track wind radii, a recent inclusion in hurricane databases. To simulate a predicted forecast, the regression model, trained over a sliding window of features, first predicts various cyclone features. Then, based on those regressed features and an additional sliding window, a classification model classifies the status of the cyclone at that point of time of the predicted features. Because of imbalances in the dataset, data augmentation techniques are explored to generate more instances of minority statuses. To effectively evaluate the model performances, the regressed errors are converted to related units, various evaluation metrics are analyzed, and the predicted trajectory on a test case is plotted.

\section{Background}
\subsection{Current Forecasting Models}
Numerical weather prediction models are computer simulations of the atmosphere that use mathematical equations to forecast future weather conditions, providing the foundation of weather forecasts \cite{NCEI_NWP2025}. Their computed numerical data assist in overall weather forecasts, requiring significant computational power and memory. The National Hurricane Center (NHC) uses statistical-dynamical methods, which are prone to truncation error, to predict the trajectory of hurricanes \cite{Alemany2019}. In addition to satellite imagery, aircraft, ship, buoy, and radar data, dynamical models, which solve physical equations of the atmosphere, and statistical models, which focus on historical cyclone data, are also generally considered \cite{NHC_ModelSummary}. 

\subsection{Cyclone Attributes}
Cyclone statuses can be classified in HURDAT form as a two-letter code indicating the type and intensity of a tropical cyclone \cite{HURDAT2_Data}. There are 9 statuses for cyclones recorded in the dataset: tropical cyclone of tropical depression intensity ($<$34 kt) (TD), tropical cyclone of tropical storm intensity (34-63 kt) (TS), tropical cyclone of hurricane intensity ($\geq$64 kt) (HU), extratropical cyclone (EX), subtropical cyclone of subtropical depression intensity ($<$34 kt) (SD), subtropical cyclone of subtropical storm intensity ($\geq$34 kt) (ST), a low that is neither a tropical, subtropical, nor an extratropical cyclone (LO), tropical wave (WV), and disturbance (DB) \cite{Landsea2015} \cite{Landsea2016}. 

The maximum wind speed of a tropical cyclone is the highest average wind speed in knots (kt) measured at a 10-meter altitude during a one minute time span, as defined by the United States National Weather Service \cite{AMS_SustainedWind}. The air pressure is correlated to intensity, but pressure-wind relationships contain some variance. The minimum pressure of a tropical cyclone is calculated by measuring the pressure at the storm’s eye and subtracting it from the surrounding ambient pressure \cite{Chavas2017}. This difference in pressure, measured in millibars (mb), offers insight to the cyclone’s intensity. A larger pressure difference, for example, generally indicates a more intense cyclone. The NHC estimates cyclone size using wind radii in four quadrants: NW, NE, SE, and SW. This radius is the largest distance from the center of the cyclone of a particular sustained surface wind speed threshold somewhere in a quadrant surrounding the center \cite{Lajoie2008}.

\section{Related Work}
Machine learning techniques can be effective for regressing cyclone trajectories and classifying cyclone conditions for improving weather forecasting accuracy. This section discusses the research of various supervised ML methods to predict locations and statuses of cyclones in related environments.
This study finds inspiration from Alemany et al. (2019), which proposed a recurrent neural network (RNN) to predict hurricane paths at 6-hour intervals \cite{Alemany2019}. To reduce truncation errors, the study implements a grid-based approach in trajectory prediction. The RNN model was able to effectively predict paths for up to 120 hours. Compared to NHC models, the RNN exhibited comparable or superior accuracy, performing with about a 50 km margin of error.

Nayak (2023) used 66 years of cyclone track data \cite{Nayak2023}, examining decision trees, random forest, naive Bayes, and SVM for storm status classification. The 2018 study reported random forest, an ensemble classifier, with the highest accuracy of 98.09\%. The dataset may have contained biases due to historical inconsistencies in cyclone recording, and although the model exhibits an excellent accuracy, other evaluation metrics were not considered.

Ibrar et al. (2025) detected and quantified tropical cyclone-related disturbances in estuaries using a Long Short-Term Memory (LSTM)-based autoencoder model, which can directly learn non-linear and temporal dependencies \cite{Ibrar2025}. The parameters considered for detection were temperature, salinity, dissolved oxygen, and turbidity. The LSTM-autoencoder was shown to effectively distinguish between natural variability and tropical cyclone-induced disturbances in estuarine water quality. However, the model provides an initial framework for disturbance predictions and needs to be applied across different climatic regions for further validation. The proposed deep learning method offers an innovative solution for anomaly detection in U.S. estuaries, and the methods are demonstrated to be functionally applicable to other uses of classifying weather conditions.

Tirone et al. (2024) integrated machine learning to classify severe thunderstorm wind reports as either severe or non-severe based in the form of probability-based classifications \cite{Tirone2024}. The study evaluated the performance of three ML models in addition to a stacked generalized linear model. The researchers also incorporated text analysis using the Correlated Topic Model (CTM) for damage descriptions. The results with the stacked generalized linear model achieved the best performance. At the Hazardous Weather Testbed Spring Forecasting Experiments (2020-2022) \cite{HWT2025}, Tirone et al. showed practical benefits for forecasting and validation with their proposed model. The models however had polarizing outputs with high and extremely low probabilities, highlighting the limitations of using objective metrics alone. This study's discussion reveals valuable examinations of non-deep learning methods with models like random forest, gradient boosting, and logistic regression. 

Arul et al. (2022) used anemometric data to detect the presence of thunderstorms using shapelet transform and random forest classifiers across North Mediterranean ports \cite{Arul2022}. Shapelet transform was a key contributor as it identifies time-series patterns based on fluctuations in the data. The tested model improved event detection accuracy by capturing distinctive wind signatures rather than relying on predefined thresholds, enabling a more automated approach to analyzing extreme wind events. However, shapelet transform can be computationally expensive, and data must be further validated with different climatic regions.

\subsection{Governing Equations}
This paper uses three governing equations during preprocessing of the dataset. As some of the ML models require scaled data values for optimal performance, certain features are normalized and standardized. The longitudinal and latitudinal points are normalized since the boundary values from -180 to 180 degrees are known. Equation \ref{eq:normalization} normalizes such points.

\begin{equation}
x' = \frac{x-min(x)}{max(x)-min(x)}
\label{eq:normalization}
\end{equation}

Maximum wind speeds can reach up to hundreds of knots, and minimum pressures can range in the thousands. Therefore, these values need to be standardized by removing the mean and scaling to unit variance. Equation \ref{eq:standardization} calculates the standard score where $\mu$ is the mean of the samples and $\sigma$ is the standard deviation of the set.

\begin{equation}
    z=\frac{x-\mu}{\sigma}
    \label{eq:standardization}
\end{equation}

The dot product is used to calculate the direction of a cyclone’s movement from one location to another, relating the movement vector and reference vector. Equation \ref{eq:angle2vectors} shows the formula for the angle between two vectors A and B.

\begin{equation}
    \theta=\arccos{(\frac{A \cdot B}{||A|| ||B||})}
    \label{eq:angle2vectors}
\end{equation}

\section{Dataset}
\subsection{Data Visualization}
The spatio-temporal dataset that the models trained and tested on, provided by the NHC of the National Oceanic and Atmospheric Administration (NOAA), contains attributes of tropical storms in the Atlantic and North Central Pacific Basins, recorded in the HURDAT2 (HURricane DATa 2nd generation) format. There are 22 labels for 49,105 rows of cyclone data in the Atlantic dataset, and in the Pacific dataset, there are again 22 labels for 26,137 rows of cyclone data. From 1851 to 2016, there are 1,814 unique recorded cyclones in the Atlantic dataset and 1,050 unique cyclones in the Pacific dataset. A cyclone’s data includes ID, name, date, time, event, status, latitude, longitude, maximum wind, minimum pressure, and directional wind radii. These features are recorded in six hour intervals throughout the cyclone’s lifespan.

\begin{figure}[H]
    \centering
    \includegraphics[width = \columnwidth]{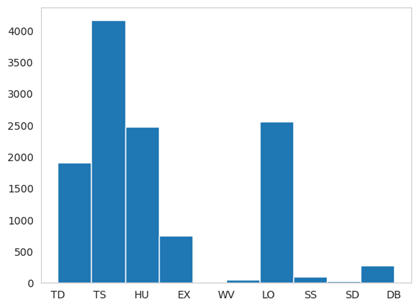}
    \caption{Distribution of cyclone status frequencies}
    \label{fig:status-distribution}
\end{figure}

Any extraneous, unidentified statuses in the dataset are removed in addition to rows that contain null values. The most abundant status in the dataset is TS with LO and then HU as the following in order. Figure \ref{fig:status-distribution} shows the distribution of cyclone statuses.

\begin{figure}[H]
    \centering
    \includegraphics[width = \columnwidth]{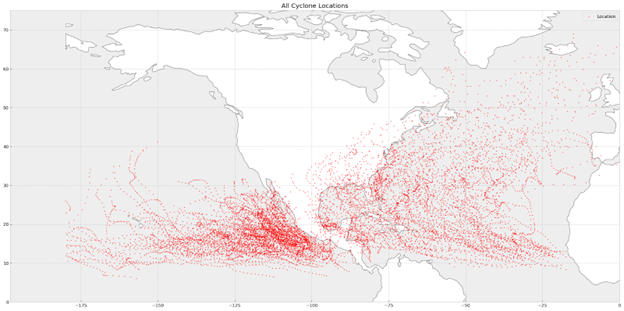}
    \caption{Cyclone trajectories in the Atlantic and Pacific Basins}
    \label{fig:cyclone-world-map}
\end{figure}

The geographic coordinates can be plotted onto a world map with the geopandas library \cite{GeoPandas2025}. Figure \ref{fig:cyclone-world-map} displays all recorded cyclone locations in the dataset in the Atlantic and North Central Pacific Basins. Each data point represents the location of the cyclone at its state at that point in time.

\subsection{Data Preprocessing}
The geographic coordinates of each cyclone datapoint is normalized into cartesian coordinates denoted by ‘x’ and ‘y’ onto a 2D grid. The month, which provides implications of temporal attributes, such as temperature or humidity, is normalized with Equation \ref{eq:normalization}. The maximum wind speeds and minimum pressures are similarly standardized with Equation \ref{eq:standardization}. Derived from the cartesian coordinates, ‘x’ and ‘y’ vector components are created that then become resultant vectors representing changes in distance covered. The angle between the resultant vector and a reference angle is then calculated using Equation \ref{eq:angle2vectors}. The reference angles used are in accordance with cardinal directions. The angle is then adjusted to be measured clockwise from north by comparing west and east angles. Thus, the final angle that measures the change in direction is achieved in radians.

Because the dataset is formatted with spatio-temporal data points, predictions are based on time series analysis. A sliding window creates data that embeds historic data in feature vectors, differentiated by their label and lag. The selected features that are included in the history window are the maximum wind speed, minimum pressure, all categories of wind radii, ‘x’ and ‘y’ coordinates, change in length, and change in direction. Generating history windows may grow to a high dimensional feature space, which may require PCA in some cases. However, a window size of five for regression and four for classification yields the lowest error and highest accuracy, respectively—thus, no dimensionality reduction is needed.

The dataset is split into training and testing by a 80/20 ratio for both regression and classification. 

\section{Methodology / Models}
\subsection{Experimental Approach}
The process is divided into a two-pronged approach. The regression model first predicts values for the maximum wind speed, minimum pressure, vector length, and vector direction based on the sliding window of size five, which contains values of the previous cartesian coordinates, maximum wind speeds, minimum pressures, wind radii, and current month. The classification models train on features including previous and current maximum wind speed, minimum pressure, vector length, vector direction, and current month. The values from the regressed vector lengths and directions are then converted into ‘x’ and ‘y’ coordinates, which are the values in addition to wind speed and pressure which the classification models then predict. In noting that each cyclone datapoint is isolated between six hour intervals, the status classification, which is based on the predicted regression values, is also a prediction of the next cyclone datapoint.

The performance of the regression model is evaluated by its mean absolute error (MAE) and the coefficient of determination (R-squared). To visualize the results and performance of the ML models, data consisting of Hurricane Katrina is extracted before training. Hurricane Katrina formed as a tropical depression over the Bahamas on August 23, 2005. Its peak intensity was categorized as a Category 5 hurricane in the Gulf of Mexico. The hurricane caused approximately \$161 billion in damage, determined as one of the costliest tropical cyclones in the Atlantic Basin \cite{BushLibrary2025}. The predicted latitudes and longitudes are plotted against the real positions of Katrina on a map of the Atlantic Basin. 

\begin{figure}[H]
    \centering
    \includegraphics[width = \columnwidth]{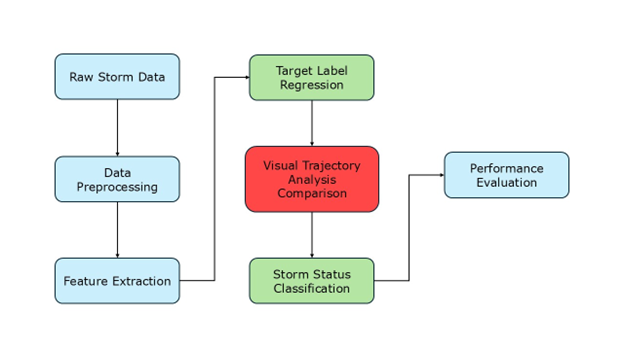}
    \caption{Flowchart of proposed process in the prediction of cyclone attributes}
    \label{fig:methodology-flowchart}
\end{figure}

The classification results are visualized with a confusion matrix and a geographic plot and evaluated by the macro and weighted averages of the precision, recall, and F1 scores in addition to the accuracy. Figure \ref{fig:methodology-flowchart} effectively shows the holistic process of this study’s approach.

\subsection{Experimental Regression Model}
This study explores the use of gradient boosting as a regression model to predict cyclone characteristics. Gradient boosting regression (GBR) is an ensemble learning technique that builds a strong predictive model by combining the outputs of multiple weak learners (typically decision trees). GBR uses a stage-wise additive approach, where each tree corrects the errors of the previous trees, optimizing the loss function using gradient descent. The residual errors are calculated and then considered for further predictions, improving overall predictions. GBR manages outliers well and is able to model complex logistic relationships, but it can be prone to overfitting if not properly tuned. GBR does not require scaled data, but because the regressed values are features for classification, and some of the classification models benefit from feature scaling, the features for regression are also scaled. 

\begin{figure}[H]
    \centering
    \includegraphics[width = \columnwidth]{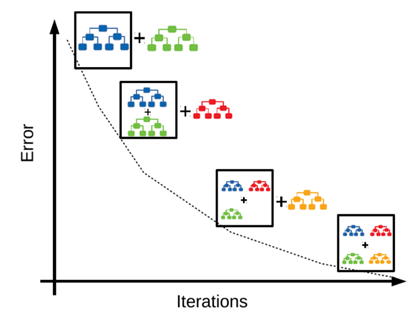}
    \caption{Process of gradient boosting regression with gradient descent}
    \label{fig:gbr-diagram}
\end{figure}

The regression model uses a history window data to predict the target values. Any regression model would likely be suitable; GBR is chosen for its simplicity and effectiveness. Figure \ref{fig:gbr-diagram} \cite{Kilari2021} shows the general process of GBR’s stage-wise additive approach and the usage of gradient descent.

\subsection{Experimental Classification Models}
Each classification model is trained with the following features: previous and current maximum wind speed, minimum pressure, ‘x’ and ‘y’ coordinates, and month. These values are derived from the GBR’s predicted values that are transformed into the appropriate representations.

\subsubsection{Random Forest (RF) Classification}
\begin{figure}[H]
    \centering
    \includegraphics[width = \columnwidth]{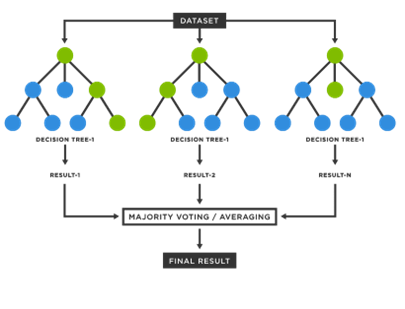}
    \caption{Process of random forest}
    \label{fig:rf-diagram}
\end{figure}

RF is another ensemble learning method, where multiple decision trees are trained on different subsets of data. A subset of features is randomly selected for each tree to prevent correlation among trees. For classification, the final prediction is made using a majority vote among the trees. RF often handles high-dimensional data well on small datasets, such as this one, and is less prone to overfitting than individual decision trees. The process of RF is displayed in Figure \ref{fig:rf-diagram} \cite{Spotfire2025}.

\subsubsection{Support Vector Machine (SVM) Classification}
\begin{figure}[H]
    \centering
    \includegraphics[width = \columnwidth]{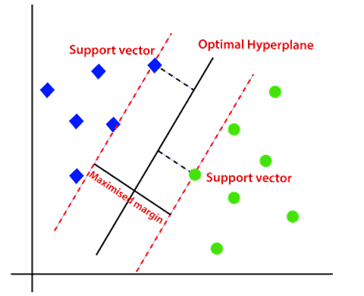}
    \caption{Example of support vector machines between two classes}
    \label{fig:svm-diagram}
\end{figure}

As a powerful supervised learning model, SVM is applicable to regression and classification tasks. The objective is to find the best decision boundary that maximizes the margin between different classes. The model constructs a hyperplane that separates individual classes in the feature space. By maximizing the margin, also known as support vectors, the model optimizes its predictions within its calculated support vector. SVM also incorporates kernel functions for non-linear, high-dimensional data. However, SVM can be computational expensive for large datasets in particular, and the chosen hyperparameters and kernels can significantly impact performance. Figure \ref{fig:svm-diagram} \cite{DataScienceLovers2025} displays an example usage of SVM.

\subsubsection{Multi-Layer Perceptron (MLP) Classification}
\begin{figure}[H]
    \centering
    \includegraphics[width = \columnwidth]{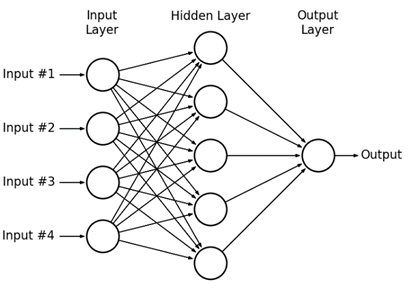}
    \caption{Process of multi-layer perceptron neural network}
    \label{fig:mlp-diagram}
\end{figure}

MLP is a type of artificial neural network that consists of multiple layers of neurons. A neural network can be effective to examine the specific contributions of various features in cyclones. All neural networks follow a similar structure. Input data is first passed through multiple layers of neutrons (input, hidden, and output layers). Each neuron applies a weighted sum followed by an activation function. The network is trained using backpropagation, where errors are propagated backward to update the weights. The optimizer minimizes the loss function using gradient descent, as seen in gradient boosting. Though in contrast, MLP also consists of hidden layers between the input and output layers. MLP requires extensive hyperparameter tuning, however, overfits data without proper regulation, and can be computationally expensive. Figure \ref{fig:mlp-diagram} \cite{Hassan2015} shows the structure of a multi-layer perceptron network.

\subsection{Synthetic Minority Oversampling Technique (SMOTE)}
\begin{figure}[H]
    \centering
    \includegraphics[width = \columnwidth]{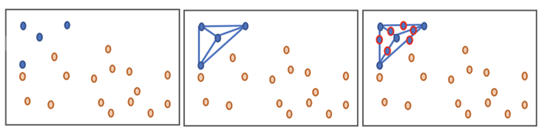}
    \caption{Example of SMOTE process on minority class}
    \label{fig:smote-example}
\end{figure}

In addition to normalizing values for regression, this study employed SMOTE to offset the imbalance of cyclone statuses in the dataset \cite{ImbalancedLearn2025}. SMOTE generates synthetic samples for the minority class by interpolating between existing minority samples and their nearest neighbors to better define classification regions. Because there are few samples of some statuses yet thousands of samples for others, SMOTE is essential to help the model learn to differentiate between statuses more accurately. Figure \ref{fig:smote-example} visually simplifies the process of SMOTE augmentation, where the SMOTE learns the structure of minority classes and generates new instances of them. 

\section{Results and Discussion}
\subsection{Regression Regression}
In determining the features that are used to classify cyclone status, the GBR model yields the following metrics seen in Table \ref{tab:GBR-metrics-table}.

\begin{table}[H]
    \centering
    \begin{tabular}{|l|l|l|l|l|}
        \hline
        Target Label & Pressure & Wind & Direction & Distance\\ \hline
        MAE & 0.121 & 0.134 & 0.372 & 0.00308\\ \hline
        R-squared error & 0.963 & 0.961 & 0.600 & 0.335\\ \hline 
    \end{tabular}
    \caption{Evaluation metric report of GBR}
    \label{tab:GBR-metrics-table}
\end{table}

\subsection{Discussion of Regression Results}
Predicted regression results exhibit excellent mean absolute errors and R-squared scores. These values represent the next features that are used to classify the cyclone. It is important to keep in mind that the ‘x’ and ‘y’ values are normalized, thus affecting length and direction, while the values of wind and pressure are standardized. Comparing the MAE for length of 0.00308 and the MAE for pressure and wind of 0.121 and 0.134, respectively, MAE for length is about three hundred times smaller than that of the other two. The disparity results from different ranges during feature scaling, where normalization scales data in the range of 0 to 1, and the range of standardization relies on the mean and standard deviation of the data.

The MAE for direction of 0.372 seems much larger than the other values, especially with regards to the MAE for length. To understand this value better, as the target values for direction are measured in radians, the MAE can be converted to an error in radians, which yields to 21.314 degrees. This error is still quite significant and may be attributed to the neglect of several other factors that influence the change in direction of a cyclone’s trajectory. Even with hyperparameter tuning, the model is unable to precisely predict changes in direction because of many other dependencies not included in the dataset that can impact a cyclone’s direction. 

The other labels can be converted into real-world measurements as well. The error in pressure comes to about 2.22 mb, in wind about 2.42 kt, and in length about 70 km. These errors are quite excellent and display competent performance with NHC methods, as most have at least 50 km in marginal error \cite{Leonardo2017}.

The model additionally displays excellent coefficients of determination of 96.3\% and 96.1\% for pressure and wind, respectively. However, direction and length received the respective low correlations of 60.0\% and 33.5\%. The R-squared score implies a weak correlation between the dependent and independent variables, which may indicate the need for more features. There may be other dependencies better suited for predicting the other target values that may not be included in the dataset. 

\subsection{Classification Results}
The three classification models–RF, SVM, and MLP–show both strengths and weaknesses in predicting cyclone status at various intervals. With hyperparameter tuning, Tables \ref{tab:RF-table}, \ref{tab:SVM-table}, and \ref{tab:MLP-table} display the macro averages and weighted averages of the precision, recall, and F1 score in addition to the overall accuracy of the prediction values by each classification model.

\begin{table}[H]
    \centering
    \begin{tabular}{|l|l|l|l|l|}
        \hline
        Average Type & Precision & Recall & F1 & Accuracy\\ \hline
        - & - & - & - & 0.93\\ \hline
        Macro Average & 0.75 & 0.78 & 0.76 & -\\ \hline 
        Weighted Average & 0.93 & 0.93 & 0.93 & - \\ \hline
    \end{tabular}
    \caption{Evaluation report of tested RF}
    \label{tab:RF-table}
\end{table}

\begin{table}[H]
    \centering
    \begin{tabular}{|l|l|l|l|l|}
        \hline
        Average Type & Precision & Recall & F1 & Accuracy\\ \hline
        - & - & - & - & 0.91\\ \hline
        Macro Average & 0.80 & 0.80 & 0.80 & -\\ \hline 
        Weighted Average & 0.91 & 0.91 & 0.90 & - \\ \hline
    \end{tabular}
    \caption{Evaluation report of tested SVM}
    \label{tab:SVM-table}
\end{table}

\begin{table}[H]
    \centering
    \begin{tabular}{|l|l|l|l|l|}
        \hline
        Average Type & Precision & Recall & F1 & Accuracy\\ \hline
        - & - & - & - & 0.90\\ \hline
        Macro Average & 0.68 & 0.79 & 0.72 & -\\ \hline 
        Weighted Average & 0.91 & 0.90 & 0.90 & - \\ \hline
    \end{tabular}
    \caption{Evaluation report of tested MLP}
    \label{tab:MLP-table}
\end{table}

\begin{figure}[H]
    \centering
    \includegraphics[width = \columnwidth]{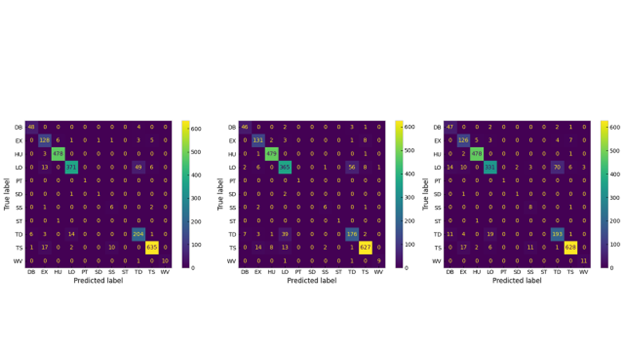}
    \caption{Confusion matrices of RF, SVM, and MLP from left to right}
    \label{fig:cfm}
\end{figure}

Confusion matrices are useful to depict any potential biases or significant misclassifications that the model performs on the test data. Figure \ref{fig:cfm} displays the confusion matrix of the RF, SVM, and MLP models from left to right.

\subsection{Discussion of Classification Results}
In comparing the three tested classification models, RF, SVM, and MLP, the model that displays the highest accuracy is RF with a score of 93\%, in contrast to SVM’s accuracy of 91\% and MLP’s of 90\%. The error can be in part due to the extremely low support values of some classes, including PT, SD, SS, ST, and WV that are close to zero in the test dataset. The low macro averaged scores in SVM and MLP indicate the model’s struggle to predict certain classes–specifically TD and LO–correctly. In fact, all models struggled to discern apart TD and LO classes as seen in Figure 14. The MLP’s  macro average precision score of 68\% is especially significant. As TD and LO statuses have similar supports, as seen in Figure 3, the issue likely arises from insufficient features by similar structures between the two classes. Although the weighted averages are higher, they are influenced by class proportions as the testing dataset is highly imbalanced.

Recall is a critical evaluation metric in the context of predicting cyclone status because certain categories are more important (e.g. hurricanes and tropical cyclones). Recall offers sensitivity for critical classes, so since false negatives (missing a dangerous cyclone) are worse than false positives (over predicting cyclones), recall should be a prioritized metric. 

Weighted averages are prioritized over macro averages because certain classes are more critical than others (e.g. HU or TS). Considering their weighted averages, RF performs better than SVM and MLP in this context because RF uses bootstrap aggregating, where each tree in the forest is trained on a random subset of the data. Fewer dominant classes thus overshadow other minorities, allowing for better generalization of minority cyclone statuses. Unlike SVM, which relies on kernels, or MLP, which requires a much larger amount of data to generalize well, RF can automatically capture complex relationships in the data, regardless of class imbalance or small datasets. Because of RF’s feature importance attribute, the model can determine which feature is most contributory for classification. Due to RF’s robustness to noise and overfitting, in predicting cyclone statuses, RF displays the strongest performance for cyclone classification. 

\subsection{Hurricane Katrina Test Case}
\begin{figure}[H]
    \centering
    \includegraphics[width = \columnwidth]{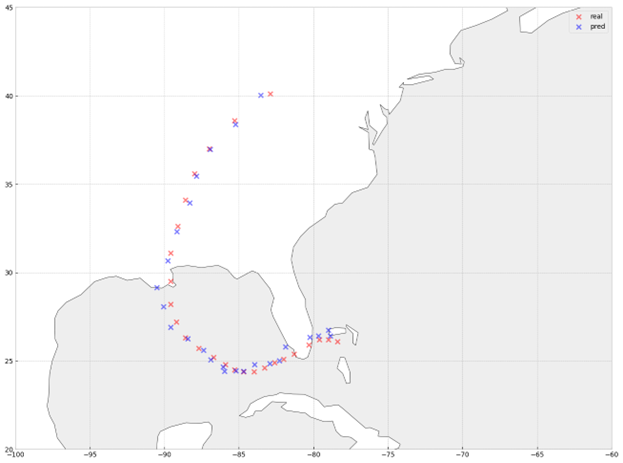}
    \caption{Visualization of regression model on Katrina}
    \label{fig:katrina-case}
\end{figure}

With the predicted values of direction and length, the trajectory of the cyclone can be tracked on a map. By converting the direction and length values to latitudinal and longitudinal points, a visual comparison of real and predicted locations can be created. Selecting Hurricane Katrina as an example test case, Figure \ref{fig:katrina-case} shows the cyclone’s real and predicted geographical points, where red points represent real values and blue points for predicted values.

The GBR model yields some improvements and some losses in predicting maximum wind, minimum pressure, changes in length, and changes in direction values, where the MAEs are shown in Table \ref{tab:GBR-Katrina-table}.

\begin{table}[H]
    \centering
    \begin{tabular}{|l|l|l|l|l|}
        \hline
       Target Label & Pressure & Wind & Direction & Distance\\ \hline
        MAE & 0.480 & 0.302 & 0.382 & 0.00259\\ \hline
        R-squared & 0.862 & 0.919 & 0.754 & 0.868\\ \hline 
    \end{tabular}
    \caption{Evaluation metric report of GBR on Katrina}
    \label{tab:GBR-Katrina-table}
\end{table}

In real-world values, the model predicts trajectories with about a 62 km error, a maximum wind speed error of 5.5 kt, and a minimum pressure error of 8.7 bar, which are still quite satisfactory results.

The classification models now predict the status of each point of Katrina. The evaluation metrics of the recall scores, as recall is the most crucial evaluation metric, are seen in Table \ref{tab:recall-katrina-table}. 

\begin{table}[H]
    \centering
    \begin{tabular}{|l|l|l|l|l|}
        \hline
       Model & Weighted Recall Average & Macro Recall Average\\ \hline
        RF & 0.90 & 0.75\\ \hline
        SVM & 0.76 & 0.38\\ \hline 
        MLP & 0.90 & 0.75\\ \hline
    \end{tabular}
    \caption{Model recall scores on Katrina}
    \label{tab:recall-katrina-table}
\end{table}

\begin{figure}[H]
    \centering
    \includegraphics[width = \columnwidth]{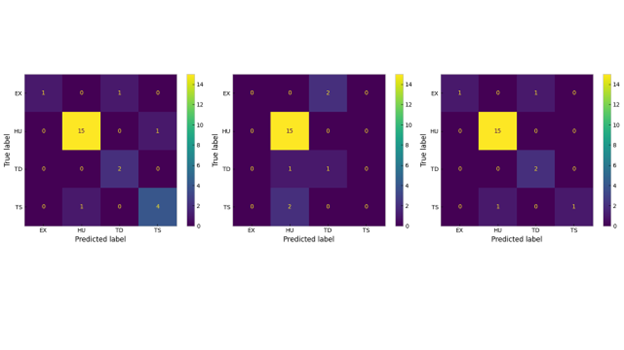}
    \caption{Confusion matrices of RF, SVM, and MLP from left to right on Katrina}
    \label{fig:cfm-katrina}
\end{figure}

The confusion matrices of each model are also seen in Figure \ref{fig:cfm-katrina} with RF, SVM, and MLP from left to right.

\begin{figure}[H]
    \centering
    \includegraphics[width = \columnwidth]{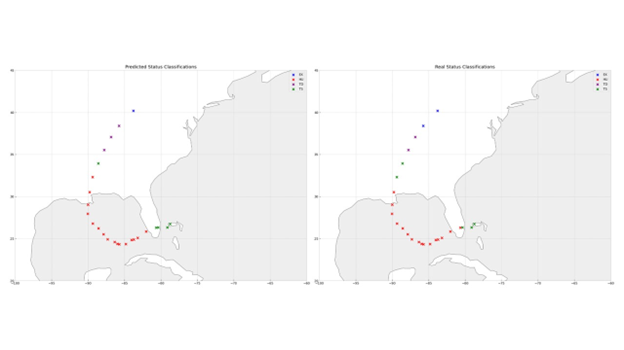}
    \caption{RF predicted statuses of Katrina}
    \label{fig:rf-katrina}
\end{figure}

It can be seen that SVM predicts poorly in comparison to RF and MLP. Although RF revealed slightly superior performance in recall during the training and testing, RF and MLP perform equally on the test case. While MLP is not as inclined to smaller datasets as RF is, if more features and instances of cyclones are considered, MLP may be able to display much better performance. RF is also able to predict labels with missing features, as seen in Figure \ref{fig:rf-katrina}, while retaining comparable if not superior results to SVM and MLP that cannot do such.

\begin{figure}[H]
    \centering
    \includegraphics[width = \columnwidth]{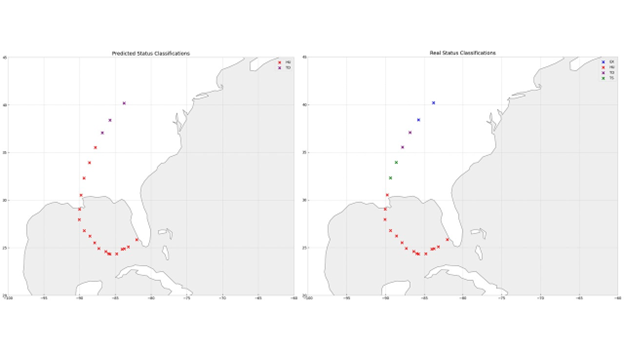}
    \caption{SVM predicted statuses of Katrina}
    \label{fig:svm-katrina}
\end{figure}

\begin{figure}[H]
    \centering
    \includegraphics[width = \columnwidth]{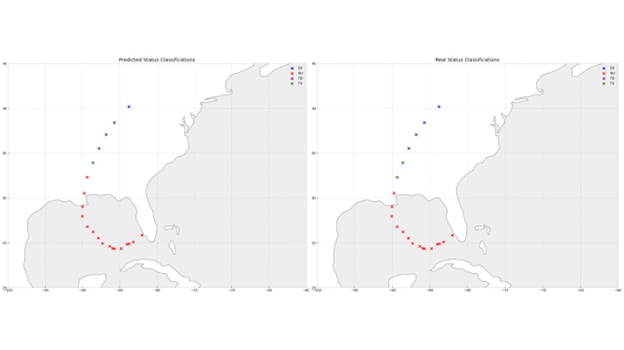}
    \caption{MLP predicted statuses of Katrina}
    \label{fig:mlp-katrina}
\end{figure}

Projected on the predicted geographic coordinates, Figure \ref{fig:rf-katrina}, \ref{fig:svm-katrina}, and \ref{fig:mlp-katrina} depict the predicted changes in status by RF, SVM, and MLP respectively to the real statuses throughout Katrina’s lifespan.

\subsection{Limitations}
As mentioned before, some of the considerable limitations of this research are the dataset subjected to few data points after preprocessing and the class imbalances. SVM optimizes for a decision boundary that maximizes margin between classes. When classes are heavily underrepresented, that margin is biased toward majority classes. Additionally with non-linear relationships, SVM may struggle to find an optimal hyperplane to separate classes effectively. Kernel tricks may aid with non-linearity but also significantly increase computational power. Deep neural networks, as with the MLP, did not perform well due to a relatively small sample size that does not include enough training examples to generalize well. The regression model also likely requires more features in order to perform with a lower MAE. As seen with predicted values of changes in direction, cyclones can be influenced by many other factors that are not incorporated in the feature training. The classification models also predict on regressed values, and because the MAE of direction is slightly concerning, classifying on predicted features may yield lower than expected results, even if performance during training is excellent.

\section{Conclusion and Future Work}
Predicting highly unpredictable cyclones is difficult due to their volatility to various factors. Nevertheless, machine learning for predicting cyclone attributes and classifying cyclones in categorical labels displays effective usage for geospatial applications. Rather than relying on physical and complex models for forecasting data, machine learning mitigates the intense computational power in such methods. In a time series analysis, the model accurately regressed maximum wind speeds, minimum pressures, and changes of length but struggled in changes of direction. Cyclones are often influenced by numerous factors, so the limited availability of features during training hinders model performance. 

RF yields the highest accuracy of 93\% out of the three models examined. Other evaluation metrics are also investigated, including precision, recall, and F1 score. The recall score is particularly significant in this study’s context, as predicting false negatives is more drastic than predicting false positives. With this metric, the sensitivity of the model is analyzed more thoroughly. The poorer performance in SVM and MLP can be explained by the model’s mechanisms and the dataset they are trained on. After preprocessing and feature extraction, the dataset becomes relatively small. Thus, SVM and MLP struggle with generalizing boundaries or patterns seen in the dataset, whereas RF excels in smaller ranges. 

Although SMOTE is employed to augment the data during training, SMOTE also struggles with generating synthetic data when there is a low support count for certain classes. The models thus cannot experience variability in low support classes, which makes such classes liable for mispredictions. The class imbalance in the dataset represents a significant limitation to classifying cyclone statuses. 
Looking forward, machine learning in regressing and classifying cyclones can be complemented with emerging technologies, such as real-time sensor networks, edge computing applications, and cloud-based processing. ML applications are becoming more and more prevalent in various industries in the world, and ML in weather forecasting introduces a potential for increased efficiency and effectiveness in saving lives and minimizing physical degradations.

\bibliographystyle{IEEEtran}
\bibliography{sample-base}
\end{document}